\documentclass{article}
\usepackage{arxiv}
\usepackage{amsmath}
\usepackage[utf8]{inputenc} 
\usepackage[T1]{fontenc}    
\usepackage{hyperref}       
\usepackage{url}            
\usepackage{booktabs}       
\usepackage{amsfonts}       
\usepackage{nicefrac}       
\usepackage{microtype}      
\usepackage{lipsum}		
\usepackage{graphicx}
\usepackage{doi}

\title{Emergence of Adaptive Circadian Rhythms in Deep Reinforcement Learning}

\author{
  \begin{tabular}{cc}
  \shortstack{Aqeel Labash\\
   Institute of Computer Science\\
   University of Tartu\\
  \texttt{aqeel.labash@gmail.com}}
 &
    \shortstack{Florian Fletzer\\
   Institute of Computer Science\\
   University of Tartu\\
   \texttt{fl.stelzer@gmail.com}} \\\\
    \shortstack{Daniel Majoral\\
   Institute of Computer Science\\
   University of Tartu\\
  \texttt{danielmajoral@gmail.com}}
 &
   \shortstack{Raul Vicente\\
   Institute of Computer Science\\
   University of Tartu\\
  \texttt{raulvicente@gmail.com}}
\end{tabular}
}
\date{}

\begin{document}
\maketitle

\begin{abstract}
Adapting to regularities of the environment is critical for biological organisms to anticipate events and plan. A prominent example is the circadian rhythm corresponding to the internalization by organisms of the $24$-hour period of the Earth's rotation. In this work, we study the emergence of circadian-like rhythms in deep reinforcement learning agents. In particular, we deployed agents in an environment with a reliable periodic variation while solving a foraging task. We systematically characterize the agent's behavior during learning and demonstrate the emergence of a rhythm that is endogenous and entrainable. Interestingly, the internal rhythm adapts to shifts in the phase of the environmental signal without any re-training. Furthermore, we show via bifurcation and phase response curve analyses how artificial neurons develop dynamics to support the internalization of the environmental rhythm. From a dynamical systems view, we demonstrate that the adaptation proceeds by the emergence of a stable periodic orbit in the neuron dynamics with a phase response that allows an optimal phase synchronisation between the agent's dynamics and the environmental rhythm.
\end{abstract}


\section{Introduction}
Circadian rhythms represent a major and well-studied adaptation of almost all terrestrial organisms to the $24$-hour rotation of the Earth \cite{minors2013circadian,panda2002circadian,vitaterna2001overview,kalsbeek2012neurobiology}. This endogenous rhythm regulates in a periodic manner the physiology of the organism, including obvious behavioral patterns such as the sleep and wakefulness cycle \cite{dunlap2004chronobiology}. At the physiological level, the circadian rhythm is best understood in Drosophila \cite{peschel2011setting}. The biochemical mechanism involves several transcription-translation feedback loops in which the transcription of genes is regulated by its protein products. These feedback loops induce the expression of so-called ``clock'' genes and protein levels to oscillate with a period of roughly $24$ hours \cite{peschel2011setting}. At the functional level, one of the key advantages of exhibiting an endogenous and entrainable rhythm is the possibility to anticipate regular events from the environment \cite{wikelski1995there}. In addition, endogenous rhythms can also synchronise interdependent physiological processes and interactions with other organisms \cite{zamm2016endogenous}.

As their biological counterparts, artificial learning agents also need to adapt to the statistical regularities of their environments. In particular, reinforcement learning agents can develop long-term strategies to explore and exploit the structure and reward signals in complex environments \cite{sutton2018reinforcement}. Intuitively,
the agent's success is explained in terms of a certain adaptation or internalization by the agent to regularities of the environment, including the environment's response to the
agent's actions. Indeed, the internalization of environmental dynamics into the agent's internal states has been related to the degree of autonomy of an agent \cite{bertschinger2008autonomy}. Thus, higher levels of autonomy indicate that the agent acts prompted by its internal state rather than being purely reactive to environmental transitions \cite{bertschinger2008autonomy,ingel2022quantifying}.

In this work, we study the specific mechanisms by which a learning agent internalizes a periodic variation in the environment. In particular, we explore how endogenous and entrainable rhythms emerge in reinforcement learning agents controlled by an artificial neural network. To this end, we deployed an agent in an environment with a reliable periodic variation while the agent learns to solve a foraging task by reinforcement learning. After characterizing the agent's behavior, we tested whether the rhythm exhibited by the agent after learning is endogenous and entrainable. Interestingly, the agent's rhythm quickly adapted to shifts in the phase of the environmental rhythm without any re-training. Using tools from dynamical systems theory, we describe how individual neurons in the network develop a stable (locally attracting) periodic orbit through a  bifurcation~\cite{strogatz2019nonlinear}. Such neural dynamics are essential to sustain the endogenous rhythm internalized by the agent. Furthermore, we compute phase response curves of these periodic orbits and explain how they help to synchronize the internalized rhythm to external inputs at an optimal phase difference. 

We remark that the model studied is not intended as a model of biological circadian rhythms. Rather the study takes inspiration from biological rhythms to select the task and assess whether and how artificial RL agents also internalize environmental regularities. It is our understanding that the internalization of environmental regularities is a general phenomenon in reinforcement learning. We study the circadian rhythm to understand in detail how a RL agent performs such internalization for the case in which the agent internalizes a simple periodic signal. That allows us to study how the internalization emerges in mechanistic terms and how the stability properties of the learned attractor endows the agent with generalization properties not foreseeable from the environment or experiences alone.

\section{Methods}

\subsection{Criteria for Circadian Rhythms}
It is generally accepted that a biological rhythm with a period of roughly $24$ hours is called circadian if the following criteria are fulfilled \cite{dunlap2004chronobiology}:

1.\,\,\textbf{Endogeneity:} A rhythm is called \textit{endogenous} if it persists without an external periodic input. That is, the rhythm must be driven by an internal mechanism of the considered organism. Specifically, circadian rhythms preserve their $24$-hour period in the absence of an external light or temperature signal that would provide a cue of the daytime.

2.\,\,\textbf{Entrainability:} A rhythm is \textit{entrainable} if it is able to adapt its phase to an external signal. Although circadian rhythms preserve a period of roughly $24$ hours even in an artificial constant environment, external cues (daylight, temperature) are necessary for readjusting the rhythm to the exact daytime. Even if a significant phase shift occurs, e.g., a change of time zones, the entrainability of circadian clocks ensures a readjustment to the environmental phase. The process of readjustment by an external signal is called \textit{entrainment}. (Note that entrainment should not be confused with training.)

3.\,\,\textbf{Temperature compensation:} The rhythm is sustained across a wide range of temperature changes. While biological processes are often accelerated by higher temperatures, circadian rhythms maintain their approximate $24$-hour period independently of the temperature of the environment.

In this work, we study an agent acting in an environment in which one variable (daylight) is periodically modulated. 
We specifically test whether the agent's rhythm has properties that correspond to the above-mentioned criteria for circadian rhythms. Our simulated environment does not contain a temperature component. Instead, we restrict ourselves to the question whether the agent's rhythmic behavior is endogenous and entrainable.

\subsection{Foraging Task and Environment}

\label{subsec:task-and-environment}

    

In our experimental setup, the environment comprises an alternating \textit{daylight signal} implementing a daytime-night-cycle. The agent's task is to collect food, which is randomly placed within a specific area, the \textit{food area}. The episode reward increases with each consumed food item. Hence, ideally the agent should try to collect as much food as possible. The placement of the food does not depend on the time of the day, i.e., the agent can find the same amount of food during daytime and night. However, at nighttime, we impose a negative reward if the agent leaves a specific safe location outside the food area. We refer to this safe location as the \textit{home location}. By choosing appropriate reward values, we ensure that the penalty for leaving the home location at night outweighs the potential reward for food collected at night. Therefore, the optimal strategy for the agent to maximize its reward is to forage in the food area at daytime and to stay at home at night. That is, the agent needs to learn to adapt its behavior to the environmental daytime-night-cycle, which is determined by a daylight binary signal.

\begin{figure}
    \centering
    \includegraphics{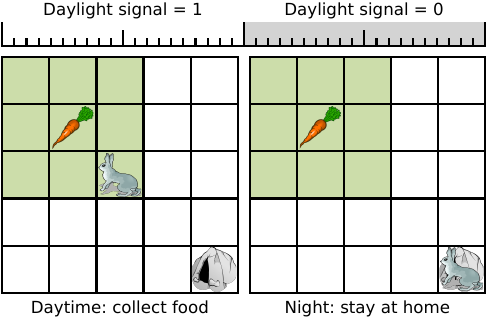}
    \caption{Environment for the foraging task. The green cells mark the food area. At the bottom right corner is the safe home location (marked by a rabbit hole), where the agent (depicted as a rabbit) must shelter at night to avoid penalties.}
    \label{fig:environment}
\end{figure}

We simulate the task and environment using the Artificial Primate Environment Simulator (APES), a customizable 2D grid world simulator~\cite{labash2018apes}. APES allows to define environments with multiple elements that interact with user-defined reward functions. For our experiments, we use a $5 \times 5$ grid world environment, illustrated in Fig.~\ref{fig:environment}. The food area consists of $3 \times 3$ grid cells indicated with a green background color. A food object is randomly placed within the food area and remains until it is collected by the agent. As soon as the food is collected, a new food item will be placed randomly within the food zone. 
The home location is the bottom right grid cell. The white grid cells represent a \textit{transit zone} separating the home location and the food area. No food will be placed in the transit zone, but the agent will still be punished for being in the transit zone at night. Only the home location is safe for the agent to shelter at night.

The time of the environment is discrete. One daytime-night-cycle consists of $40$ time steps. Accordingly, the daylight signal is periodic with a period of $40$ time steps. It has the value $1$ (daytime) for the first $20$ time steps of a daytime-night-cycle and the value $0$ (night) for the following $20$ time steps.
We specifically designed the foraging task such that the agent needs to anticipate the progression of time, i.e., the phase of the daylight signal. The agent perceives the current state of the daylight signal, i.e., it knows at every time step whether it is currently daytime or night. However, the current state of the daylight signal does not provide information about its phase. That is, solely based on the current state, it is impossible to determine how many time steps are left until the next change from daytime to night, or vice versa. However, the agent needs at least four time steps to return from the food area to the home location. Therefore, for earning a high reward, the agent must leave the food area a few time steps before the night begins. Thus, the agent needs to develop the ability to anticipate the onset of the night to avoid penalty scores.

We call the information that the agent receives at the time step $t$ an \textit{observation} and denote it by $o_t$. Besides the daylight signal, the observation $o_t$ comprises the current location of the agent, the agent's orientation, and the location where the food is currently placed (Fig. \ref{fig:network-architecture}). Since the daylight signal is periodic, the state $s_t$ of the environment is entirely described by the observation and the phase of the daylight signal. In other words, the environment is a partially observable process, where the phase of the daylight signal is a hidden variable~\cite{Majeed2018,Hausknecht2015}.

For each time step, the agent performs one of five possible actions: moving up, down, right, left, or standing still.
The agent earns a reward of $+1$ for each consumed food object, and is penalized with a value of $-2.5$ for each time step spent outside the home location at night.
\subsection{Architecture and Training}

\label{sec:architecture}
We train an agent to perform the foraging task by deep reinforcement learning (DRL)~\cite{sutton2018reinforcement,thorndike1911animal,schultz1997neural, mnih2015human} using a dueling Q-network~\cite{Wang2016duelingDQN}\footnote{Code at \href{https://github.com/aqeel13932/MN_project}{https://github.com/aqeel13932/MN\_project}}. The network estimates so-called Q-values of state-actions pairs to enable the agent to chose the best action for the current state.
As described above, we differentiate between state $s$, which completely describes the environment, and observation $o$, which is the part of the state that is perceived by the agent.
The network input at each time step $t$ is the observation $o_t$ containing the spatial information from the environment (location of the agent and food) and the current state of the daylight signal. For optimal decisions, however, we actually need to consider the complete state $s$: the information about the progression of the day (remaining time steps until the end of the daytime) is relevant for the agent to decide when to leave the food area to return home.
Despite this information is not contained in $o_t$ (only in $s_t$), it can be extracted from the \textit{history of observations} $h_t = \{ o_t, o_{t-1}, o_{t-2}, \ldots \}$. Therefore, we equip our network with an LSTM layer~\cite{Hochreiter1997,Bakker2001,Hausknecht2015} to represent information from past inputs in its internal state.

We use experience replay to achieve stable training and an $\varepsilon$-greedy policy for exploration.
We train the network for $37500$ training episodes, each consisting of $160$ time steps (four full days).

Details of the network architecture are in Appendix~\ref{sec:appendix-architecture}. For an explanation of Q-learning with dueling Q-networks, we refer to Appendix~\ref{sec:appendix-Q-network-details}. For a details about training process and an overview of the used hyperparameters, we refer to Appendix~\ref{sec:appendix-training-details}.
The evolution of the reward during training is shown in Fig.~\ref{fig:training} in Appendix~\ref{sec:learning-the-foraging-task}.

\section{Results}
\begin{figure*}
    \centering
    \includegraphics{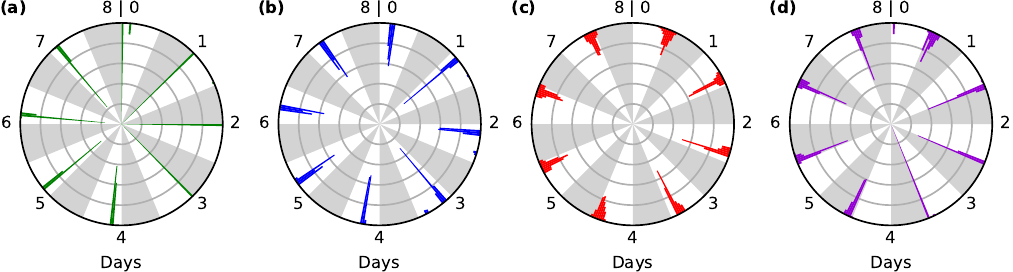}
    \caption{Timing of the agent entering and leaving the home location and food area. The angular plots show histograms of the agent's timing for eight full-day cycles. The following events are shown: (a) the agent leaves the home location, (b) the agent enters the food area, (c) the agent leaves the food area, (d) the agent returns to the home location. The grey grid lines mark event probabilities of $0.2$, $0.4$, $0.6$, etc. Daytime and night are indicated by white and grey areas. The histogram values were obtained from $1000$ test runs.    
    }
    \label{fig:roseplot}
\end{figure*}
\subsection{The Agent's Behavior}
\label{subsec:emergence}

As described in Sec.~\ref{subsec:task-and-environment}, the grid world environment comprises three distinct areas: first, the food area, where the agent can gain reward by collecting food; second, the home location, where the agent is protected from receiving penalty scores at night; and third, the transit zone, which must be crossed by the agent when moving from the home location to the food area or vice versa. 
To move from the food area to the home location, the agent needs at least four time steps. That is, to avoid a penalty at night, the agent must plan at daytime to reach home ideally exactly at the $21$st time step of the day (first time step of the night).

To evaluate the agent's behavior, we characterize how it navigates the environment by the timing of its most salient actions: leaving the home location, entering the food area, leaving the food area, and entering the home location. We trained a randomly initialized model and performed $1000$ test runs for which we captured the time steps of these events.
The results are shown in Fig.~\ref{fig:roseplot}: for each day of the test phase, we indicate the frequency of each event type by colored histograms.
The green histograms in panel a) show at which time steps the agent leaves the home location. During the first four days of the test runs, the agent typically leaves home at the first time step of the daytime, which is ideal for maximizing the reward. For the remaining four days, the agent leaves the home location a few time steps later (.e.g. during day 8 the agent leaves the home area at the fourth time step). This can be explained by the fact that our training episodes include only four days, whereas the test runs comprise eight days. The network must generalize from the training data to make suitable decisions after the fourth day. Indeed, as we note below, the average LSTM activation decreases considerably during the night of the fourth day. This is likely due to the lack of pressure from the training (which  consisted only of 4 days) to maintain any particular activation range for the LSTM right before the episode would end (during training the agent does not need to anticipate any subsequent event after day 4). Nevertheless, the agent actions and LSTM activations continue with a daily regularity from day 5 onwards. The agent's behavior reveals that this generalization is not perfect but sufficient to obtain near maximal reward. 
The time points at which the agent enters the food location are shown by blue histograms in panel b). As expected, this happens a few time steps after leaving home.
The red histograms in panel c) show the time points at which the agent leaves the food area. The agent has to make this decision in anticipation of the approaching night to enter the home location on time and avoid a large penalty. The purple histograms in panel d) show when the agent reaches the home location after leaving the food area. For all eight days, the agent arrives almost always on time or with a delay of at most four time steps. This shows that the generalization from the shorter training episodes (lasting only four days) works well for learning to predict the onset of the night.

\subsection{Testing the Rhythm Endogeneity}
\label{subsec:endogeneity}

One of the main characteristics of circadian rhythms is their endogeneity, i.e., their property to maintain their period even in the absence of an external periodic drive signal.
During training, we use a periodic daylight signal, which has the value $1$ at daytime, i.e., from time step $1$ to $20$ of each day cycle, and the value $0$ at night (time steps $21$ to $40$).
To demonstrate the endogeneity of the observed circadian behavioral rhythm (as opposed to just being a sequence of reactions to the external daylight cues), we need to consider cases where the daylight signal is clamped to a constant value, $1$ or $0$, to model permanent daytime or permanent night, respectively. In the following, we describe the results of such tests under constant conditions. Further, we perform a bifurcation analysis of the activity of the LSTM units in the network.

\begin{figure}[t]
    \centering
    \includegraphics{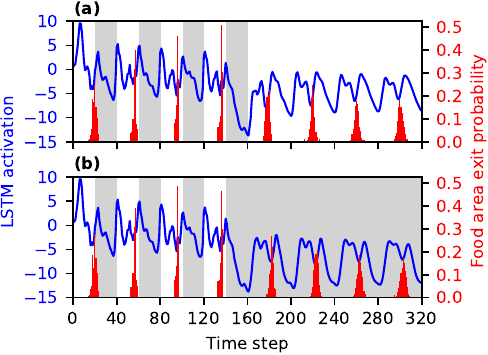}
    \caption{Agent behavior and LSTM activation for a clamped daylight signal: (a) permanent daytime and (b) permanent night after the fourth day. Shown are average values over $1000$ test simulations. Daytime and night are represented by white and grey areas. The average activation of the LSTM neurons is plotted in blue. The activation pattern is shaped differently after day $4$, but it retains its period.
    The red bars are histograms counting the agent's exits of the food area at the respective time step.}
    \label{fig:perm}
\end{figure} 

\subsubsection{Tests During Constant Conditions}
To confirm the endogeneity of the agent's rhythm, we use test runs in which after certain time the daylight signal is clamped to a constant value. Hence, we apply the usual periodic daylight signal for four day cycles ($160$ time steps) to ensure that the agent's rhythm is present and phase-adjusted to the environmental time. For the remainder of the test run (time steps 160 to 320) a constant daylight signal was applied (either constant daytime or constant night). Figure~\ref{fig:perm} shows the mean activation of the LSTM neurons and the timing of the agent's behavior for $1000$ test runs under these conditions. The activation pattern shows that the rhythm persists with a period of roughly one day ($40$ time steps). We can conclude that the observed rhythm is endogenous: for days $5$ to $8$, the oscillation of the LSTM activations and the behavior of the agent are not forced externally. During this phase, the oscillation is purely the result of the dynamical properties of the trained neural network. In other words, the LSTM layer internalized the environmental periodic rhythm.

\begin{figure}[t]
	\centering
	\includegraphics{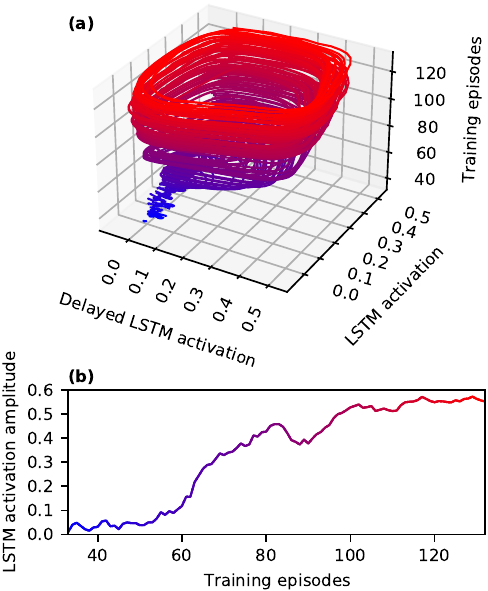}
	\caption{
    Panel (a) shows the activation of an arbitrarily chosen LSTM neuron for a constant daylight signal (permanent daytime) plotted against itself delayed by $1/4$ period. Shown are data obtained by one test run after each training episode from $33$ (blue) to $132$ (red). Panel (b) shows the amplitude of the neuron's activation.
    }
	\label{fig:bifurcation}
\end{figure}

\subsubsection{Bifurcation analysis}
Without training, the LSTM activations do not oscillate if we apply a constant daylight signal. The internal rhythm arises during the initial training phase. That is, the step-wise parameter change during training causes the system to undergo a bifurcation (a sudden qualitative change in a system behavior due to a small change in the system parameters). This bifurcation is illustrated by Fig.~\ref{fig:bifurcation}a, which shows the activation of one arbitrarily chosen LSTM neuron plotted against the delayed activation (delayed by $10$ time steps, i.e., $1/4$ period) of the same neuron during the training episodes $33$ to $132$.
Although the plot depicts the activation state of a single neuron, other neurons in the LSTM layer act in a qualitatively equivalent way.
The plot shows that the neuron's activation state remains approximately zero for the initial training episodes. However, after roughly $55$ episodes, we observe a rhythmic behavior, which indicates the onset of a stable periodic orbit in the system's dynamics.
Figure~\ref{fig:bifurcation}b depicts the amplitude (here defined as the difference between maximum and minimum) of the neuron's activation. Between episode $55$ and $100$, we observe cycles with increasing amplitude. After episode $100$, the amplitude remains nearly constant. The period of the neuron's activation is always roughly one day ($40$ time steps). The fact that the amplitude grows from zero after the system crosses the bifurcation, whereas the frequency immediately jumps to a positive value, suggests that the observed stable periodic orbit  emerges as a result of a supercritical Neimark-Sacker bifurcation~\cite{kuznetsov1998elements}.

\begin{figure}[t]
	\centering
	\includegraphics{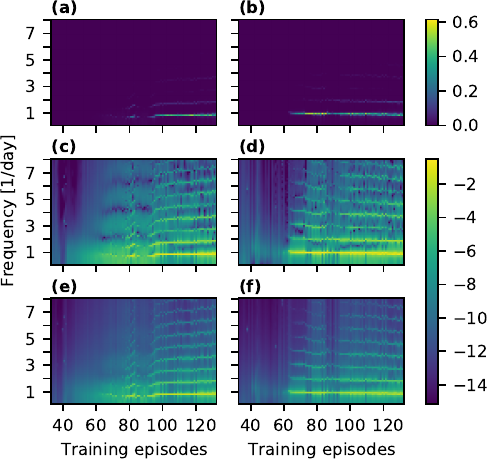}
	\caption{
	Power spectra of the LSTM activation for the initial training phase. The top row contains power spectra of one arbitrary LSTM neuron for (a) permanent daytime and (b) permanent night. Panels (c) and (d) show the logarithm of the same spectral data. Applying the logarithm emphasizes subtle frequency pattern such as the higher order resonances or the pattern at early episodes. The bottom row shows the logarithm of the mean power spectra of all $128$ LSTM neurons for (e) permanent daytime and (f) permanent night.
	All spectral data were calculated based on a single test simulation.
    }
	\label{fig:power-spectrum-init}
\end{figure}

The above observations are further supported by the spectrograms shown in Fig.~\ref{fig:power-spectrum-init}. The upper row of the figure contains spectrograms of the activation of an arbitrarily chosen LSTM neuron for constant daylight signal: panel (a) illustrates the permanent daytime case, and panel (b) the permanent night case. The color represents the power spectral density of the neuron's activation. The spectrograms exhibit a frequency peak near $1/\textrm{day}$ beginning after $60$ to $80$ training episodes, which validates our above observation of a frequency jump at the bifurcation point. Panels (c) and (d) show the logarithm of the power spectral density of the same neuron as in panels (a) and (b). Plotting the logarithmic values enables us to see subtle frequency peaks: higher order resonances of the base frequency show up, and, in particular for panel (c), it is revealed that the frequency peak at $1/\textrm{day}$ emerges at an earlier period than visible in panel (a). Finally, panels (e) and (f) in Fig.~\ref{fig:power-spectrum-init} show the logarithm of the mean of the power spectra of all $128$ LSTM neurons for permanent daytime and permanent night, respectively. The single neuron spectra shown in panels (a) to (d) are very similar to the average spectra shown in panels (e) and (f). In fact, all LSTM neurons reveal a similar spectral pattern, and hence a similar frequency content.

Additionally, in Appendix~\ref{sec:appendix-long-term-frequency}, we plot spectrograms (Fig.~\ref{fig:power-spectrum-long}) for the whole training procedure of $37500$ episodes. It is revealed that in the course of the training, the system may undergo further bifurcations. Once learned, the internalized rhythm is in fact persistent for the whole training phase and can always be observed if the daylight signal is clamped to $1$ (permanent daytime) during the test runs. However, the rhythm can switch between being active and inactive if the daylight signal is clamped to $0$ (permanent night) as shown in Fig.~\ref{fig:power-spectrum-long}b.

\subsection{Tests for Rhythm Entrainability}

Entrainability is the ability of an oscillating system to synchronize to an oscillating input signal or environment. It is one of the defining properties of circadian rhythms and ensures that the internal circadian clock is continuously adjusted to the environmental clock time. Moreover, it enables the circadian rhythm to readjust to sudden phase shifts of the environmental rhythm. A prominent example is the ability of humans and other biological organisms to adapt to time differences when travelling across time zones. 

\subsubsection{Jet lag experiments}
\label{subsubsec:Jet-lag-experiments}

\begin{figure}[t]
    \centering
    \includegraphics{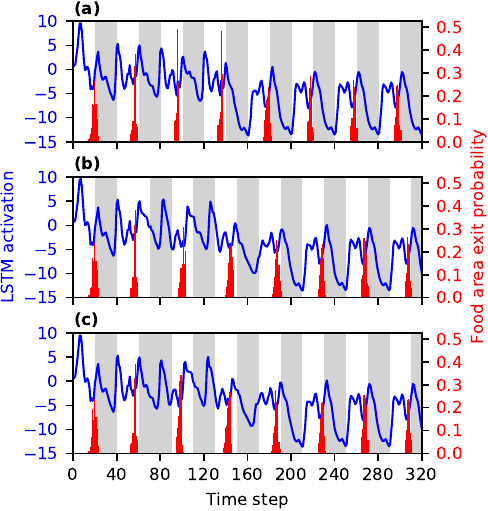}
    \caption{Effect of a daylight signal phase shift on the agent's behavior and the LSTM activation. Shown are (a) a base case without phase shift, and two dephased cases, where (b) the second daytime period, or (c) the second night is extended by $10$ time steps. All values are averages of $1000$ test simulations. Daytime and night are represented by areas with white or grey background. The average activation of the LSTM neurons is plotted in blue. The red bars are histograms counting the agent's exits of the food area at the respective time step. Both the LSTM activation and the behavioral pattern adapt to the phase shift within three days.}
    \label{fig:timing}
\end{figure}

As with all the experiments in this paper we did not re-train the model. To confirm and study the entrainability of the network's rhythm, we simulated time zone shifts by altering the length of a single daytime period or night during test runs with the trained model. Figure~\ref{fig:timing} shows how the time shifts affect the average activation pattern of the LSTM layer and the timing of the agent's behavior. We altered the length of the second day by extending the daytime (Fig.~\ref{fig:timing}b) or the night (Fig.~\ref{fig:timing}c) by $50$ percent. For comparison, we show the results for the unaltered case (Fig.~\ref{fig:timing}a). For the cases with extended daytime or night, we observe a ``jet lag'' effect on day $3$: the agent exits the food area earlier than necessary. This jet lag effect is slightly stronger for the extended night case. In both cases, the agent progressively adapts its food area exit time to the new environmental time on day $4$ and day $5$, which indicates that the controlling neural network is re-synchronising its internal clock to the changed environmental time. On day $6$ and later, the agent's internal rhythm is again synchronised with the environment. Further experiments with phase perturbations, including complete reversal of daytime and night, are described in section D.3 in the Appendix. This shows that the neural network dynamics is able to compensate within three days for strong shifts of the environmental clock.

\subsubsection{Phase response curve analysis}
The effect of environmental light on the circadian rhythms of humans has been studied by measuring the change (phase response) of the human rhythm as a reaction to light exposure at different times of the day \cite{minors1991human}. To obtain these measurements, test persons stayed within an environment with controlled light conditions for a couple of days and were exposed to bright light during certain time periods. The light exposure resulted into a phase shift of the circadian clock of the test persons, which could be determined by measuring their body-temperature curve throughout the day.
These observations can be visualized with phase response curve (PRC), which plots the phase shift in reaction to a light exposure (the phase response) against the phase when the light exposure occurred.
The PRC studies in humans provide several insights. Light at the evening or early night leads to a negative phase response, i.e., a delay of the circadian clock. On the contrary, light at the late night and morning causes a positive phase response, i.e., the circadian clock is advanced. Consequently, the PRC in an average human crosses the x-axis at two points: at night with a positive slope, and at daytime with a negative slope. The phase of the circadian rhythm is in a stable equilibrium if the time of the light exposure is correctly adjusted with the negative slope zero-crossing of the PRC. This observation describes the regulation of the human circadian clock by light from a dynamical systems viewpoint.

Motivated by these studies of the human PRC, we plotted PRCs for the LSTM layer of our trained neural network. We ran a series of test simulations with the usual alternating daylight signal on the first four days, and a constant daylight signal for the subsequent days, i.e., either permanent daytime or permanent night. On day $5$, we inverted the daylight signal for one time step, i.e., we applied a daylight pulse if the signal was permanent night and vice versa. Then we measured the resulting phase shifts on day $6$. The initial period of four days with periodic daylight signal sets the neural network's inner clock. Therefore, we can interpret the measured phase shifts at given times of day $5$ as the network's reaction to light (or darkness) at the corresponding phase of its internal clock.

\begin{figure}[t]
	\centering
	\includegraphics{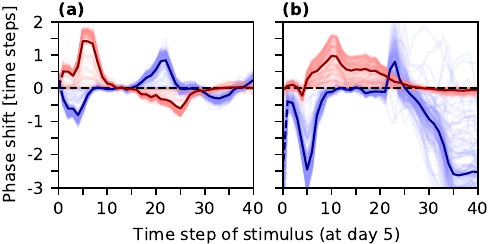}
	\caption{
    Average phase response to light exposure (red) and darkness (blue). Panels (a) and (b) show the PRCs for two independently initialized and trained models. The light red and light blue curves are the PRCs for the $128$ individual neurons of the LSTM layer obtained by averaging over $200$ test runs. The dark red and dark blue curves show the mean phase response of all LSTM neurons.
    }
	\label{fig:prc}
\end{figure} 

Figure~\ref{fig:prc} shows the PRCs of the LSTM neurons reacting to a light impulse (red curves) and to darkness (blue curves) for two randomly initialized and independently trained models (panels (a) and (b)).
In case (a), the agent reacts with a positive phase shift to light and a negative phase shift to darkness during the early morning time around time step $1$. This corresponds to a phase advance for more light in the early morning and a phase delay for darkness at this time. At evening time (near time step $20$) this pattern is inverted and can be interpreted equivalently. In panel (a), the model response to light (red curve) is similar to the human PRC for light exposure. In case (b), the most significant phase shift occurs if light is absent in the morning (late nighttime and early daytime). This suggests that the corresponding trained model resets its inner clock mainly during morning time.
The PRCs shown in panels (a) and (b) differ significantly, which indicates different learned phase adjustment mechanisms. This difference between the trained models indicates that entrainment can be realized by different strategies and that the neural network is able to learn more than one of these strategies.

\subsection{Robustness of the emergence of circadian-like rhythms}
So far the emergence of a stable periodic orbit representing the internalisation of an environmental rhythm has been studied for a particular architecture and type of recurrent layer (LSTM). Hence, a natural question concerns the generality of the phenomenon. To explore this question we ran a series of variations on the network architecture, type of recurrent layer, and learning algorithm. In particular, we considered variations in the optimization algorithm (SGD and RMSprop), weight regularization (L1 + L2 norms), weight initialization (He normal), type of recurrent layer (vanilla RNN and GRU), as well as the width of the recurrent layer (32 and 96 neurons) and fully connected layer (8 units). 

For each case, we tested whether the variations also resulted in the endogeneity and entrainability of a rhythm emerging in the recurrent layer of the network. Appendix \ref{sec:robustness} shows the results for the constant conditions and jet-lag experiments for all the variations. Overall, the emergence and entrainability of the internal rhythm were robust to most of the variations explored.

We also noted that different seeds might lead to different strategies, an effect well known in reinforcement learning. While it is not possible to fully explore and characterize the conditions under which one type of solution or another emerges, the results indicate that the circadian-like properties of the emergent rhythm are not particular to a specific set of parameters or network architecture but rather they occur in a wide range of conditions.

\section{Related Work}
Studying circadian rhythms was a natural choice and inspiration since they constitute a clear and almost universal internalization of a simple environmental rhythm, and for which different techniques for assessing the internalization of the rhythm were available (tests for endogeneity and entrainment could be borrowed from the biological literature). In particular, we studied how the system dynamics supports the internalization and its adaptation properties. More generally, the environmental periodic signal that we considered represents a specific case of non-stationarity in the environmental properties (cyclo-stationarity). How an optimal agent adapts to this and other non-stationarities (including those due to the adaptation of other agents \cite{tampuu2017multiagent}) is a question of theoretical and practical interest for continual learning in RL \cite{khetarpal2012towards}. Interestingly, the internalization and stability properties of the periodic orbit by the agent's neurons  endowed the agent with robustness to perturbations (jet-lag experiments) not foreseeable from the training experiences.

The internalization of environmental correlations has been related to the degree of autonomy of an agent \cite{bertschinger2008autonomy}.
In \cite{ingel2022quantifying} a partial information decomposition was used to quantify the degree of internalization in tasks with a Markovian dynamics. The estimated index was a global measure of internalization that cannot distinguish what dynamics is being internalized nor its mechanisms. In the present work, we explicitly demonstrated the internalization of an environmental rhythm by a reinforcement learning agent, and how this happened via a bifurcation in the LSTM units. This bifurcation endowed the network with a stable periodic orbit with phase entrainable dynamics.

Bifurcations of random RNNs and their node synchronisation have been studied in \cite{marquez2018dynamical} with an emphasis on how they impact their computational properties. For an attractor view on RNNs training, see also \cite{ribeiro2020beyond}.

The dynamical systems view of neural networks has been influential in the development of neural ordinary differential equations (ODE's) and its variants. In particular, dynamical stability of vanilla RNNs and its relation to vanishing and exploding gradients was addressed in \cite{chang2019antisymmetricrnn}, where a neutral stability condition was imposed by restricting weight matrices to be anti-symmetric. Numerical results demonstrated that this condition improved the training and generalization of the network in several tasks. Early work on considering RNNs as dynamical systems and how their stability can impact the training can be found in \cite{pascanu2013difficulty}. In our work, we did not impose any stability condition, rather we characterized how the phase stability of an emergent attractor explained the adaptability of the agent to the perturbations of the external rhythm that were never experienced during the training episodes. 

To our knowledge, studies of how artificial neural networks develop attractors and how the attractor characteristics relate to the training and generalisation properties have been conducted in the supervised  setting. For example, in \cite{ansuini2019intrinsic} the authors observed that smaller intrinsic dimensions of representations in the final layers of vision networks correlated with higher accuracy in a classification task. Line attractor dynamics (continuous attractors with one direction of neutral stability) have been reported in sentiment analysis networks \cite{maheswaranathan2019reverse}. 

The ability of LSTM units to distinguish precise timing in temporal patterns is well known and it was demonstrated in early work \cite{gers2002learning}. Here we were interested in the type of solution adopted by the agent. We note that the agent trained in our study could potentially have developed a simple event-driven mechanism that counted time steps triggered by the external daylight transitions to solve the task without internalising any rhythm. This was not the case in our study where a rhythm of appropriate periodicity was clearly internalized. While a simple counter triggered by the environment transitions is one of the optimal solutions for the training of the foraging task, the presence of a sustained rhythm in the agent under constant environmental conditions revealed that the internalization of the rhythm was the actual solution adopted during learning. For other network internalizations (or tasks), it is possible that simple counting mechanisms emerge as possible solutions. What are the exact factors that determine the emergence of one or another type of solution is a matter of further investigation. 

Models of circadian rhythms abound in the mathematical biology literature. These models often consist of coupled differential equations describing concentration of molecules, gene expression levels, or multi-cellular changes \cite{asgari2019mathematical}. No learning or reinforcement mechanisms are included in these studies where the model parameters are fixed or scanned. For a recent application of artificial circadian rhythms in robotics, see \cite{o2021entraining}.

\section{Discussion}

We have investigated the emergence of circadian-like rhythms in deep learning agents trained by reinforcement in a foraging task. The results show that a reinforcement learning agent equipped with LSTM units can internalize an external rhythm and use it to anticipate and exploit an environmental regularity. In particular, the timing of the agent's actions was critically controlled by the internalized rhythm. We conducted extensive experiments to determine the properties of the agent's rhythm. Tests under constant conditions and jet lag experiments confirmed that the rhythm was endogenous and entrainable in a similar way as circadian rhythms exhibited by biological organisms. Furthermore, bifurcation and phase response curve analyses were conducted to characterize the emergence of the rhythm and its synchronization properties. We observed the emergence of a stable periodic orbit in the LSTM dynamics via a bifurcation as the training progressed. Since the periodic orbit emerges with a smoothly increasing amplitude and an instantaneous jump of the frequency, we conjecture a supercritical Neimark-Sacker bifurcation.
The phase response curves illustrate how the phase of the agent's internal clock is dynamically attracted by the phase of the environmental rhythm via phase-dependent reactions to the daylight signal.
This stability property ensures that the agent can adapt to phase shifts in the environment. Interestingly, the phase stability emerged, although the agent has not experienced phase perturbations during training (which always consisted of four regular daytime-night-cycles). Moreover, the phase response curves reveal that the agent is not limited to learning one specific strategy. A comparison of two independently trained models shows significant differences in the phase response of the periodic orbits. This observation raises the question whether the observed periodic orbit may stem from different types of bifurcations.

Our results are in line with the view that learning agents can develop long-term strategies by internalizing correlations in the environment dynamics and the agent-environment interactions. Planning ahead often requires a simulation or unfolding in time of the dynamics to be predicted. In the case that we studied, such an internalization led to the emergence of a periodic trajectory of LSTM units that enabled the agent to anticipate the environmental dynamics. 

As mentioned above, we can understand the adaption and internalization of the circadian-like rhythm by the agent as the effect of a bifurcation, i.e., a parameter change in a dynamical system (LSTM units) which causes topologically different trajectories and attractors as the training progresses. The neural network controller of the agent developed a periodic orbit with stability properties that endowed the agent with an endogenous and entrainable rhythm. More generally, this observation raises the question whether agents trained in different tasks and environments also benefit from developing attractors whose topology and stability support appropriate computations and policies. From this perspective, successful learning is directly related to changing parameters of the model to induce the appropriate bifurcations and attractors to support the representation and computations of appropriate variables. This is already a successful research direction in neuroscience \cite{khona2022attractor,chaudhuri2019intrinsic} that could be transferred to the study of how artificial learning agents represent and process information by exploiting attractor dynamics. 

\section*{Acknowledgements}
We are thankful to Jaan Aru, Tambet Matiisen, Meelis Kull, and three anonymous
reviewers for constructive comments on the manuscript.

This research was supported by the Estonian Research Council Grant PRG1604, the European Union’s Horizon 2020 Research and Innovation Programme under Grant Agreement No. 952060 (Trust AI), the Estonian
Centre of Excellence in IT (EXCITE) Project Number TK148, and the Project CardioStressCI (ERA-CVDJTC2020-015) from the European Union’s ERA-CVD Joint Transnational Call 2020.

\bibliography{references}

\begin{thebibliography}{10}

\bibitem{minors2013circadian}
David~S Minors and James~M Waterhouse.
\newblock {\em Circadian rhythms and the human}.
\newblock Butterworth-Heinemann, 2013.

\bibitem{panda2002circadian}
Satchidananda Panda, John~B Hogenesch, and Steve~A Kay.
\newblock Circadian rhythms from flies to human.
\newblock {\em Nature}, 417(6886):329--335, 2002.

\bibitem{vitaterna2001overview}
Martha~Hotz Vitaterna, Joseph~S Takahashi, and Fred~W Turek.
\newblock Overview of circadian rhythms.
\newblock {\em Alcohol Research \& Health}, 25(2):85, 2001.

\bibitem{kalsbeek2012neurobiology}
Andries Kalsbeek, Martha Merrow, Till Roenneberg, and Russell~G Foster.
\newblock {\em The neurobiology of circadian timing}.
\newblock Elsevier, 2012.

\bibitem{dunlap2004chronobiology}
Jay~C Dunlap, Jennifer~J Loros, and Patricia~J DeCoursey.
\newblock {\em Chronobiology: biological timekeeping.}
\newblock Sinauer Associates, 2004.

\bibitem{peschel2011setting}
Nicolai Peschel and Charlotte Helfrich-F{\"o}rster.
\newblock Setting the clock--by nature: circadian rhythm in the fruitfly
  drosophila melanogaster.
\newblock {\em FEBS letters}, 585(10):1435--1442, 2011.

\bibitem{wikelski1995there}
Martin Wikelski and Michaela Hau.
\newblock Is there an endogenous tidal foraging rhythm in marine iguanas?
\newblock {\em Journal of Biological Rhythms}, 10(4):335--350, 1995.

\bibitem{zamm2016endogenous}
Anna Zamm, Chelsea Wellman, and Caroline Palmer.
\newblock Endogenous rhythms influence interpersonal synchrony.
\newblock {\em Journal of Experimental Psychology: Human Perception and
  Performance}, 42(5):611, 2016.

\bibitem{sutton2018reinforcement}
Richard~S Sutton and Andrew~G Barto.
\newblock {\em Reinforcement learning: An introduction}.
\newblock MIT press, 2018.

\bibitem{bertschinger2008autonomy}
Nils Bertschinger, Eckehard Olbrich, Nihat Ay, and J{\"u}rgen Jost.
\newblock Autonomy: An information theoretic perspective.
\newblock {\em Biosystems}, 91(2):331--345, 2008.

\bibitem{ingel2022quantifying}
Anti Ingel, Abdullah Makkeh, Oriol Corcoll, and Raul Vicente.
\newblock Quantifying reinforcement-learning agent’s autonomy, reliance on
  memory and internalisation of the environment.
\newblock {\em Entropy}, 24(3):401, 2022.

\bibitem{strogatz2019nonlinear}
S.~Strogatz.
\newblock {\em Nonlinear Dynamics and Chaos: With Applications to Physics,
  Biology, Chemistry, and Engineering}.
\newblock Chapman \& Hall book. CRC Press, 2019.

\bibitem{labash2018apes}
Aqeel Labash, Ardi Tampuu, Tambet Matiisen, Jaan Aru, and Raul Vicente.
\newblock Apes: a python toolbox for simulating reinforcement learning
  environments.
\newblock {\em arXiv preprint arXiv:1808.10692}, 2018.

\bibitem{Majeed2018}
Sultan~Javed Majeed and Marcus Hutter.
\newblock On q-learning convergence for non-markov decision processes.
\newblock In {\em Proceedings of the Twenty-Seventh International Joint
  Conference on Artificial Intelligence, {IJCAI-18}}, pages 2546--2552.
  International Joint Conferences on Artificial Intelligence Organization, 7
  2018.

\bibitem{Hausknecht2015}
Matthew Hausknecht and Peter Stone.
\newblock Deep recurrent q-learning for partially observable mdps.
\newblock In {\em 2015 aaai fall symposium series}, 2015.

\bibitem{thorndike1911animal}
Edward~Lee Thorndike.
\newblock {\em Animal intelligence: Experimental studies}.
\newblock Macmillan, 1911.

\bibitem{schultz1997neural}
Wolfram Schultz, Peter Dayan, and P~Read Montague.
\newblock A neural substrate of prediction and reward.
\newblock {\em Science}, 275(5306):1593--1599, 1997.

\bibitem{mnih2015human}
Volodymyr Mnih, Koray Kavukcuoglu, David Silver, Andrei~A Rusu, Joel Veness,
  Marc~G Bellemare, Alex Graves, Martin Riedmiller, Andreas~K Fidjeland, Georg
  Ostrovski, et~al.
\newblock Human-level control through deep reinforcement learning.
\newblock {\em nature}, 518(7540):529--533, 2015.

\bibitem{Wang2016duelingDQN}
Ziyu Wang, Tom Schaul, Matteo Hessel, Hado Hasselt, Marc Lanctot, and Nando
  Freitas.
\newblock Dueling network architectures for deep reinforcement learning.
\newblock In Maria~Florina Balcan and Kilian~Q. Weinberger, editors, {\em
  Proceedings of The 33rd International Conference on Machine Learning},
  volume~48 of {\em Proceedings of Machine Learning Research}, pages
  1995--2003, New York, New York, USA, 20--22 Jun 2016. PMLR.

\bibitem{Hochreiter1997}
Sepp Hochreiter and Jürgen Schmidhuber.
\newblock {Long Short-Term Memory}.
\newblock {\em Neural Computation}, 9(8):1735--1780, 11 1997.

\bibitem{Bakker2001}
Bram Bakker.
\newblock Reinforcement learning with long short-term memory.
\newblock In T.~Dietterich, S.~Becker, and Z.~Ghahramani, editors, {\em
  Advances in Neural Information Processing Systems}, volume~14. MIT Press,
  2001.

\bibitem{kuznetsov1998elements}
Yuri~A Kuznetsov, Iu~A Kuznetsov, and Y~Kuznetsov.
\newblock {\em Elements of applied bifurcation theory}, volume 112.
\newblock Springer, 1998.

\bibitem{minors1991human}
David~S Minors, James~M Waterhouse, and Anna Wirz-Justice.
\newblock A human phase-response curve to light.
\newblock {\em Neuroscience letters}, 133(1):36--40, 1991.

\bibitem{tampuu2017multiagent}
Ardi Tampuu, Tambet Matiisen, Dorian Kodelja, Ilya Kuzovkin, Kristjan Korjus,
  Juhan Aru, Jaan Aru, and Raul Vicente.
\newblock Multiagent cooperation and competition with deep reinforcement
  learning.
\newblock {\em PloS one}, 12(4):e0172395, 2017.

\bibitem{khetarpal2012towards}
Khimya Khetarpal, Matthew Riemer, Irina Rish, and Doina Precup.
\newblock Towards continual reinforcement learning: A review and perspectives;
  2020.
\newblock {\em URL https://arxiv. org/abs/2012}, 13490, 2012.

\bibitem{marquez2018dynamical}
Bicky~A Marquez, Laurent Larger, Maxime Jacquot, Yanne~K Chembo, and Daniel
  Brunner.
\newblock Dynamical complexity and computation in recurrent neural networks
  beyond their fixed point.
\newblock {\em Scientific Reports}, 8(1):1--9, 2018.

\bibitem{ribeiro2020beyond}
Ant{\^o}nio~H Ribeiro, Koen Tiels, Luis~A Aguirre, and Thomas Sch{\"o}n.
\newblock Beyond exploding and vanishing gradients: analysing rnn training
  using attractors and smoothness.
\newblock In {\em International Conference on Artificial Intelligence and
  Statistics}, pages 2370--2380. PMLR, 2020.

\bibitem{chang2019antisymmetricrnn}
Bo~Chang, Minmin Chen, Eldad Haber, and Ed~H Chi.
\newblock Antisymmetricrnn: A dynamical system view on recurrent neural
  networks.
\newblock {\em arXiv preprint arXiv:1902.09689}, 2019.

\bibitem{pascanu2013difficulty}
Razvan Pascanu, Tomas Mikolov, and Yoshua Bengio.
\newblock On the difficulty of training recurrent neural networks.
\newblock In {\em International conference on machine learning}, pages
  1310--1318. PMLR, 2013.

\bibitem{ansuini2019intrinsic}
Alessio Ansuini, Alessandro Laio, Jakob~H Macke, and Davide Zoccolan.
\newblock Intrinsic dimension of data representations in deep neural networks.
\newblock {\em Advances in Neural Information Processing Systems}, 32, 2019.

\bibitem{maheswaranathan2019reverse}
Niru Maheswaranathan, Alex Williams, Matthew Golub, Surya Ganguli, and David
  Sussillo.
\newblock Reverse engineering recurrent networks for sentiment classification
  reveals line attractor dynamics.
\newblock {\em Advances in neural information processing systems}, 32, 2019.

\bibitem{gers2002learning}
Felix~A Gers, Nicol~N Schraudolph, and J{\"u}rgen Schmidhuber.
\newblock Learning precise timing with lstm recurrent networks.
\newblock {\em Journal of machine learning research}, 3(Aug):115--143, 2002.

\bibitem{asgari2019mathematical}
Ameneh Asgari-Targhi and Elizabeth~B Klerman.
\newblock Mathematical modeling of circadian rhythms.
\newblock {\em Wiley Interdisciplinary Reviews: Systems Biology and Medicine},
  11(2):e1439, 2019.

\bibitem{o2021entraining}
Matthew~Joseph O'Brien.
\newblock {\em Entraining a Robot to its Environment with an Artificial
  Circadian System}.
\newblock PhD thesis, Georgia Institute of Technology, 2021.

\bibitem{khona2022attractor}
Mikail Khona and Ila Fiete.
\newblock Attractor and integrator networks in the brain.
\newblock {\em Nature reviews neuroscience}, 23:744--766, 2022.

\bibitem{chaudhuri2019intrinsic}
Rishidev Chaudhuri, Berk Ger{\c{c}}ek, Biraj Pandey, Adrien Peyrache, and Ila
  Fiete.
\newblock The intrinsic attractor manifold and population dynamics of a
  canonical cognitive circuit across waking and sleep.
\newblock {\em Nature neuroscience}, 22(9):1512--1520, 2019.

\bibitem{labash2020perspective}
Aqeel Labash, Jaan Aru, Tambet Matiisen, Ardi Tampuu, and Raul Vicente.
\newblock Perspective taking in deep reinforcement learning agents.
\newblock {\em Frontiers in Computational Neuroscience}, 14, 2020.

\bibitem{chollet2015keras}
Fran\c{c}ois Chollet et~al.
\newblock Keras.
\newblock \url{https://keras.io}, 2015.

\end{thebibliography}
\bibliographystyle{unsrt}

\newpage
\appendix
\renewcommand\thefigure{\thesection.\arabic{figure}}    
\renewcommand{\thetable}{A\arabic{table}}
\setcounter{figure}{0} 
\setcounter{table}{0}
\onecolumn

\section{Architecture}

\label{sec:appendix-architecture}


We train an agent to perform the foraging task, described in Sec.~\ref{subsec:task-and-environment}, using deep reinforcement learning (DRL)~\cite{sutton2018reinforcement,thorndike1911animal,schultz1997neural}. That is, the agent is controlled by a deep neural network that decides which action should be performed at which state to obtain the largest possible reward.
A standard approach for DRL is Q-learning~\cite{mnih2015human}, where a deep neural network learns to estimate so called Q-values, i.e., a weighted sum of future rewards (called the discounted return) that we can expect for certain actions at a given state. We refer to this type of neural network as deep Q-network (DQN). Based on the estimates provided by the DQN, the agent can decide which action it takes.

For our numerical experiments, we use a dueling Q-network~\cite{Wang2016duelingDQN} , which is an enhanced version of the original DQN. Instead of estimating the Q-values directly, a dueling Q-network provides two distinct estimates. Firstly, it approximates the state value, which is the expected discounted return for a given state. Secondly, it approximates advantage values, which indicate how much better the expected return is for a specific action at the given state relative to the other possible actions.

The Q-value of an action depends on the state $s_t$ of the environment. In particular during daytime, to determine which action is the most promising, we need to know the remaining time until the night begins. This information is part of the state $s_t$ but not provided by the observation $o_t$ at time $t$. However, it can be extracted from the \textit{history of observations} $h_t = \{ o_t, o_{t-1}, o_{t-2}, \ldots \}$. Therefore, we chose a network architecture with an internal state that allows us to represent information from past inputs: we equip our network with an LSTM layer~\cite{Hochreiter1997,Bakker2001,Hausknecht2015}.

\begin{figure}
\centering
\includegraphics{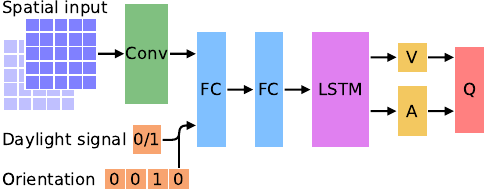}
\caption{Network architecture. The network input consist of a binary tensor that represents the location of the agent and the food, the scalar-valued daylight signal, and a one-hot vector encoding the orientation of the agent. The spatial information fed to a convolutional layer whose output is flattened and concatenated with the agent's orientation vector and the daylight signal. This is followed by two fully connected layers and an LSTM layer, after which the network is split into two branches: branch $V$ estimates the state value, branch $A$ estimates the advantage values for all possible actions. Both estimates are combined to obtain the network's output estimating the Q-values.
}
\label{fig:network-architecture}
\end{figure}

The network architecture of our dueling Q-network is illustrated in Fig.~\ref{fig:network-architecture}. The network comprises two input layers and multiple hidden layers. The input of the network at time step $t$ is the observation $o_t$. The first input layer is a $5 \times 5 \times 2$ binary tensor, encoding the location of the agent and food objects in our $5 \times 5$ grid world. The tensor has the value one at the current location of the agent (or the food, respectively) and zero values elsewhere. The second input layer has five nodes representing the daylight signal and the agent's orientation. The daylight signal has the value one at daytime and zero at night. The agent's orientation is encoded in a one-hot vector.
The first hidden layer is a ReLU-activated convolutional layer that receives its input from the first input layer. That is, the convolutional layer processes the binary tensor providing the spatial information of the environment (the agent and the food location). It has six output channels of size $5\times 5$, which are generated using kernels of size $3\times 3$. The output of the convolutional layer is flattened and merged with the daylight signal and the orientation vector. The convolutional layer is followed by two fully connected layers with $32$ nodes each, and one LSTM layer with $128$ cells. For these fully connected layers and the LSTM layer we use the hyperbolic tangent as activation function. The LSTM layer is followed by two branches containing small fully connected layers that estimate the state value and the advantage values. Naturally, the state value layer contains only one neuron (because it estimates a one-dimensional quantity). The advantage value layer contains five neurons, one for each possible action. For both layers, we use ReLU activation. As final step, both estimates are combined to obtain the Q-value estimates, which can be used to govern the agent's behavior.

\section{Dueling Q-networks}
\label{sec:appendix-Q-network-details}

In Appendix~\ref{sec:appendix-architecture}, we describe the architecture of the duelling Q-network with an LSTM layer that controls the agent. Based on the history of observations $h_t = \{ o_t, o_{t-1}, o_{t-2}, \ldots \}$, the network estimates the Q-values, i.e., the expected discounted return for state-action-pairs $(s, a)$. Thereby, the Q-value estimates are obtained from separate estimates of the state value and the advantage values for each action.

The discounted return at time step $t$ is formally defined as $R_t = \sum_{\tau = t}^{\infty} \gamma^{\tau -t} r_\tau$, where $r_\tau$ is the reward obtained at time step $\tau$ and $\gamma \in [0,1]$ is a discount factor.
The return that we can expect from a given state and action strongly depends on the behavior of the agent. Therefore, Q-value, state value, and advantage value are defined based on a policy $\pi = \mathbb{P}(a \mid h)$, which describes the probability that the agent takes action $a$ for a history of observations $h$. The policy $\pi$ can be stochastic or deterministic (where all probability values are either one or zero). The Q-value for state $s$ and action $a$ under policy $\pi$ is defined by
\begin{equation}\label{eq:q-function}
    Q^\pi (s,a) = \mathbb{E}_{\pi}[R_t \mid s_t = s, a_t = a],
\end{equation}
where $s_t$ and $a_t$ are the state and action at time step $t$. Note that in our case the Q-value must be defined as the expected value of the return even if the policy is deterministic because our environment is stochastic.
The state value is defined as the expected Q-value for the considered state:
\begin{equation}\label{eq:v-function}
    V^\pi (s) = \mathbb{E}_{a\sim \pi} [Q^\pi (s,a)].
\end{equation}
The advantage value of taking a particular action $a$ at a state $s$ is then defined as the difference between Q-value and state value:
\begin{equation}\label{eq:a-function}
    A^\pi (s,a) = Q^\pi (s,a) - V^\pi (s).
\end{equation}
For distinction, we denote the functions $Q^\pi$, $V^\pi$, and $A^\pi$ defined in Eqs.~\eqref{eq:q-function}--\eqref{eq:a-function} with superscript $\pi$, whereas we denote the corresponding Q-network and its state value and advantage value branches (which approximate these functions) by $Q$, $V$, and $A$.

When we train the network $Q$, we fit the network to the actual Q-function $Q^\pi$. The state value branch $V$ and the advantage value branch $A$ are not directly fitted to $V^\pi$ and $A^\pi$ but trained via backpropagation. In order to achieve that these branches are indirectly fitted to $V^\pi$ and $A^\pi$, the network output $Q$ must be implemented in a suitable way. We cannot simply use the sum of $V$ and $A$ for the network output, even though this would comply with Eq.~\eqref{eq:a-function}. If we did, the output would remain unchanged if we increase $V$ by an arbitrary constant and decrease $A$ by the same constant. Hence, the training process would not force $V$ and $A$ to $V^\pi$ and $A^\pi$. Therefore, we implement the network output via
\begin{equation}\label{eq:q-output}
    Q(h,a) = V(h) +\left[ A(h,a) - \max_{a'} A(h,a') \right].
\end{equation}
As a result, $V$ approximates $V^\pi$, where $\pi$ is the policy to chose the best action, and the term $A(h,a) - \max_{a'} A(h,a')$ approximates the advantage of action $a$ in comparison to the best action.

The original publication~\cite{Wang2016duelingDQN} that introduced dueling Q-networks proposed Eq.~\eqref{eq:q-output} and an alternative equation using the mean instead of the max operator. We tested both versions and found that implementing the network output with Eq.~\eqref{eq:q-output} works better in our case.

\section{Training Details}
\label{sec:appendix-training-details}
We followed the training procedure of~\cite{mnih2015human}. We train our Q-network by recursively exploiting the relation between the Q-value of a state-action-pair $(s, a)$ and the Q-values for the consequence state $s'$ which we obtain by taking action $a$ at state $s$. For this recursive training approach, Q-value estimates for the consequence state are employed to define a target for the Q-network $Q(h,a; \theta)$, where $\theta$ is the parameter set and $h$ is the history of observations. The Q-value estimates for the consequence state are obtained using a separate target network with parameter set $\theta^-$. The target for $Q(h, a; \theta)$ is defined as
\begin{equation}
    y = r + \gamma \max_{a'} Q (h', a', \theta^-),
\end{equation}
where $r$ is the reward for the action $a$ taken at the current time step, $h'$ is the new history of observations at the next time step (after taking action $a$), and $\gamma$ is a discount factor for future rewards. The target value $y$ is only an approximation of the actual Q-value, but it is indeed good enough for training the network $Q(h,a;\theta )$ if we use an adequate behavioral policy and update rule for $\theta^-$.
For fitting $Q(h,a;\theta )$ to $y$, we use the loss function
\begin{equation} \label{eq:loss}
    L(\theta) = \mathbb{E} \left[ (y-Q(h,a ; \theta))^2 \right],
\end{equation}
which we optimize by gradient descent.
The strategy of using a separate target network $Q(h',a'; \theta^-)$ to compute the targets~$y$ was developed to stabilize the training procedure~\cite{mnih2015human}. In the original publications that introduced the target network approach~\cite{mnih2015human} and dueling Q-networks~\cite{Wang2016duelingDQN}, the parameters $\theta^-$ of the target network were updated periodically. The parameter set $\theta^-$ was fixed for a certain number of training steps while the online network's parameters $\theta$ were progressively updated. After each period, the target network's parameters were set to $\theta^- = \theta$. For our network training, we use a modified approach. We define the target network's parameters as an exponential moving average of the online weights and update them at every time step via
\begin{equation}
    \theta^- = \beta \theta + (1-\beta )\theta^-,
\end{equation}
where $\beta = 0.001$.

For further improvement of the training stability, we use experience replay~\cite{mnih2015human}. An experience consists of the history of observations $h$, the chosen action $a$, the reward $r$ obtained at the considered time step, and the new history of observations $h'$ after taking the action $a$. We store the observations $o$, the chosen actions $a$, and the rewards $r$ for the $1000$ most recent training episodes in a replay memory. Thus, we can retrieve all experiences $(h,a,r,h')$ of these $1000$ episodes from the replay memory. For updating the network parameters $\theta$, we compute the gradient of the loss function~\eqref{eq:loss} by averaging over $16$ experiences that were randomly drawn from the replay memory (for each update step).
Hence, the formula for the loss function can be written as
\begin{equation} \label{eq:loss-replay}
    L(\theta) = \mathbb{E}_{(h,a,r,h') \sim \mathcal{D}} \left[ (y-Q(h,a ; \theta))^2 \right],
\end{equation}
where $\mathcal{D}$ is the replay memory.

In order to find a good strategy for solving the foraging task, we need a policy to produce exploratory actions. Therefore, we apply an $\varepsilon$-greedy policy. That is, with probability $\varepsilon$, the agent takes a randomly chosen action, and otherwise, the agent takes the action $a^\ast = \mathrm{argmax}_a Q(h,a;\theta)$ which maximizes the current Q-value estimate.

We train our model for $37500$ training episodes, each of which consists of $160$ time steps. That is, we train for a total number of $6$ million time steps. The daylight signal has a period of $40$ time steps. For the first $20$ steps within this period, the daylight signal has the value $1$ (daytime), for the remaining $20$ steps it has the value $0$ (night). That is, each training episode corresponds to four simulated day cycles.
For each time step, we perform four training steps, i.e., four updates of the parameter set $\theta$ of the online network. For this, we draw four samples consisting of $16$ training episodes, and perform one $\theta$-update step for each sample. The first $32$ training episodes are only used to fill the experience replay memory. No parameter updates are performed during these $32$ initial episodes.
The network is trained using the learning rate $\eta = 0.001$. The exploration parameter $\varepsilon$ is linearly annealed from $1$ to $0.1$ for the first $75\%$ of the training and constantly $0.1$ for the remaining training steps.

\begin{table}
\caption{Hyperparameters of the neural network architecture and training.}
\label{tab:hyperparameters}
\vskip 0.15in
\begin{center}
\begin{small}
\begin{sc}
\begin{tabular}{ll}
\toprule
Hyperparameter & Value \\
\midrule
    Kernel size of the convolutional layer & $3\times 3$ \\
    Number of output channels of the convolutional layer & $6$ \\
    Size of the fully connected layers & $32$ \\
    Number of LSTM cells & $128$ \\
    Reward for collecting food & $1$ \\
    Penalty for each time step spent outside the home location at night & $-2.5$ \\
    Discount factor for the discounted return & $\gamma = 0.99$ \\
    Number of training episodes & $37500$ \\
    Number of time steps per episode & $160$ \\
    Number of training steps (gradient descent steps) per time step & $4$ \\
    Replay memory size (number of training episodes in memory) & $1000$ \\
    Training sample size (number of episodes sampled from replay memory) & $16$ \\
    Learning rate & $\eta = 0.001$ \\
    Exploration parameter & $\varepsilon = 1.0$ to $0.1$  \\
    Update rate for the target network & $\beta = 0.001$ \\
\bottomrule
\end{tabular}
\end{sc}
\end{small}
\end{center}
\vskip -0.1in
\end{table}

The hyperparameters that we used for our network and training are the same as in~\cite{labash2020perspective}. The hyperparameters are summarized in Table~\ref{tab:hyperparameters}. The network was implemented using the \texttt{Keras 2.1.5} library~\cite{chollet2015keras}.

\section{Additional Results}
\subsection{Learning the Foraging Task}
\label{sec:learning-the-foraging-task}

We evaluated the agent's ability to maximize the reward as the training progresses. The agent obtains a reward of $+1$ for each consumed food item and a penalty of $-2.5$ for each time step spent outside the home location at night. That is, to maximize the reward, the agent needs to collect as many food items as possible at daytime and it needs to stay at the home location at night.

\begin{figure}
\centering
\includegraphics{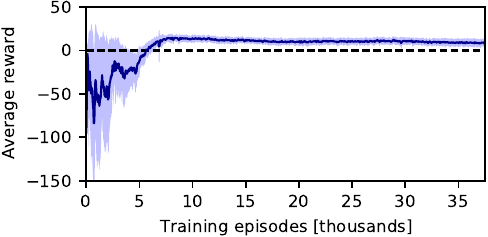}
\caption{Reward improvement during the training progress. The solid blue line shows the average reward gained in test runs of the agent during the training process. The data is obtained by averaging over seven randomly initialized models and over a window of $11$ test episodes using the central moving average. The standard deviation of the reward is indicated by the light blue area.}
\label{fig:training}
\end{figure}

We trained seven randomly initialized models for $37500$ training episodes. After every tenth training episode, we performed a test run and recorded the obtained reward.
The results are shown in Fig.~\ref{fig:training}: after roughly $5000$ training episodes, the agent is still unable to gain a positive reward. Moreover, we can observe a strong fluctuation of the obtained reward during this initial training phase, which is reflected in a large standard deviation. After this initial phase, the agent's performance improves significantly. The largest average reward is achieved after approximately $8000$ training episodes. With further training, the average reward decreases slightly, but stabilizes at a level near the optimum.

\subsection{Bistable Endogeneity Pattern During Training}

\label{sec:appendix-long-term-frequency}

In Sec.~\ref{subsec:endogeneity}, we demonstrate the endogeneity of the learned rhythm by applying constant daylight signals. For both permanent daytime and permanent night, we observe a rhythmic behavior with a period of approximately one day. Spectrograms of the LSTM activation during the initial training phase suggest that the endogeneity emerges through a supercritical Neimark-Sacker bifurcation.

\begin{figure}
	\centering
	\includegraphics{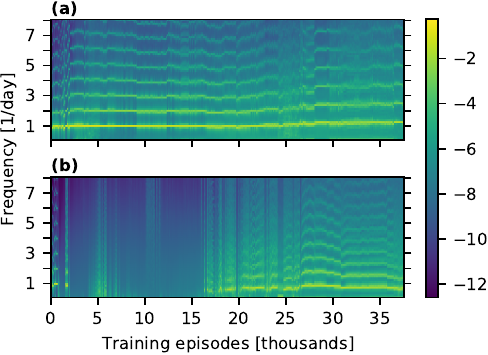}
	\caption{
	Power spectra of the LSTM activation for the whole training procedure. Shown is the logarithm of the power spectrum of one arbitrary (but typical) LSTM neuron for (a) permanent daytime and (b) permanent night. The spectrograms show that the network training establishes a steady cyclic behavior for permanent daytime (daylight signal clamped to one). For permanent night, it is not guaranteed that the cyclic behavior is preserved by the network training. 
	In both cases the spectral data were calculated based on a single test simulation.
    }
	\label{fig:power-spectrum-long}
\end{figure}

Inspecting the frequency content beyond the initial training phase reveals that the internalized rhythm persists for the whole training phase, but is not necessarily active at nighttime. Figure~\ref{fig:power-spectrum-long} depicts the log power spectrum of an arbitrary neuron over the full training period consisting of $37500$ training episodes with a resolution of $100$ episodes; i.e., test runs were performed after every $100$th training episodes to record the data that are shown in this figure. Two test cases are presented: (a) permanent daytime and (b) permanent night. It is revealed that the LSTM passes further bifurcation points during the training. While the cyclic behavior remains present for permanent daytime, it disappears and reappears multiple times for permanent night. A possible explanation for this observation is that anticipating the onset of the night is essential for obtaining a high reward. If the agent is not able to return to the home location before nighttime, it will receive high penalty scores. Therefore, it is necessary to keep track of the current time step at daytime, which the network does using its internal rhythm. In contrast, at nighttime the LSTM activity is not absolutely necessary because the agent must stay inactive until the last time step of the night, i.e., the agent can simply react to the start of the daylight signal to go to the food area.

\subsection{Simulations With Extreme Changes in Phase}
\label{sec:Extra-simulations}
\begin{figure}[t]
    \centering
    \includegraphics[trim={12mm 12mm 12mm 12mm},clip]{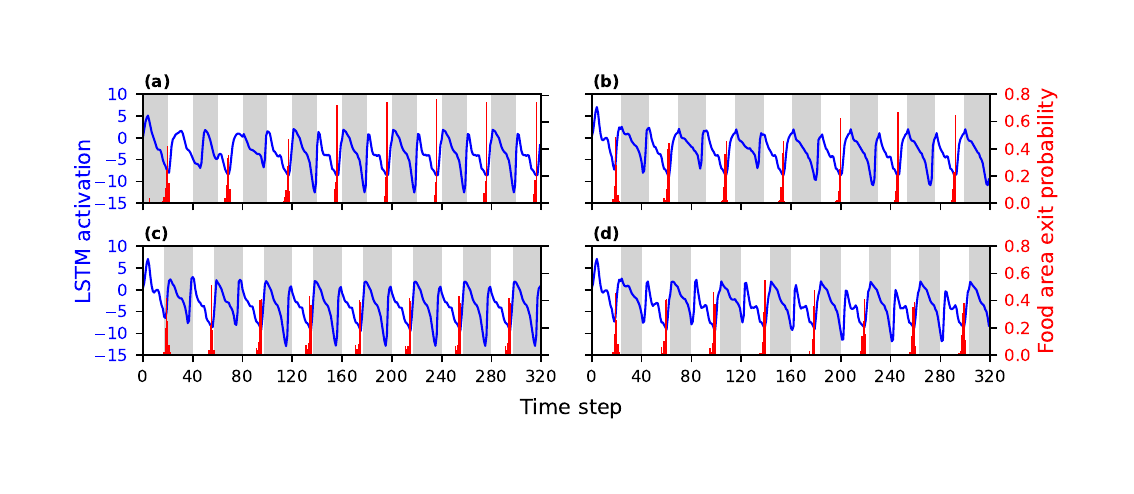}
    \caption{Effect of a daylight signal phase shift on the agent's behavior and the LSTM activation. Panel (a) shows the case where the daytime and night were reversed. In (b) we introduce a longer daytime-night cycle consisting of 23 time steps of daytime and 23 time steps of night (23,23). In (c) and (d) we permanently change the ratio between daytime and night respectively to become (17,23) in (c) and (23,17) in (d).}
    \label{fig:timing11}
\end{figure}

In Sec \ref{subsubsec:Jet-lag-experiments} we explored 2 experiments where the daytime or night of a single day period were extended. In Fig. \ref{fig:timing11} we show more extreme simulations were the model showed similar adaptation. In Fig. \ref{fig:timing11} (a), we switched the daytime and night which is the opposite signal received during training. The results shows that the agent adapted to the changes after 4 daytime-night-cycle. In Fig. \ref{fig:timing11} (b), , which changes the total period of the daytime-night-cycle to 46 time steps. The change in frequency of the external cue is matched by the frequency of the LSTM activation. That is, the LSTM is able to lock to the new frequency of forcing. However, the locking to the modified external frequency results in a additional phase difference that makes the agent to leave the food area too early for obtaining an optimal reward. The effect of of permanently changing ratio between daytime and night while keeping the same overall period of 40 time steps is shown in panels c) and d). The LSTMs guiding the behavior of the agent are able to adapt to such changes and control the agent to exit the food area in a near optimal time.

\subsection{Robustness against variations of network architecture, type of recurrent layer, and optimization algorithm}
\label{sec:robustness}

To determine the generality of the emergence of the circadian-like rhythms, we considered
variations in the optimization algorithm (SGD
and RMSprop), weights regularization (L1 + L2 norms), weights initialization (He normal), type of recurrent layer (vanilla RNN and GRU), as well as the width of the recurrent layer
(32 and 96 neurons) and fully connected layer (8 units). 

\begin{figure}[t]
    \centering
    \includegraphics[trim={12mm 12mm 12mm 12mm},clip,width=\textwidth]{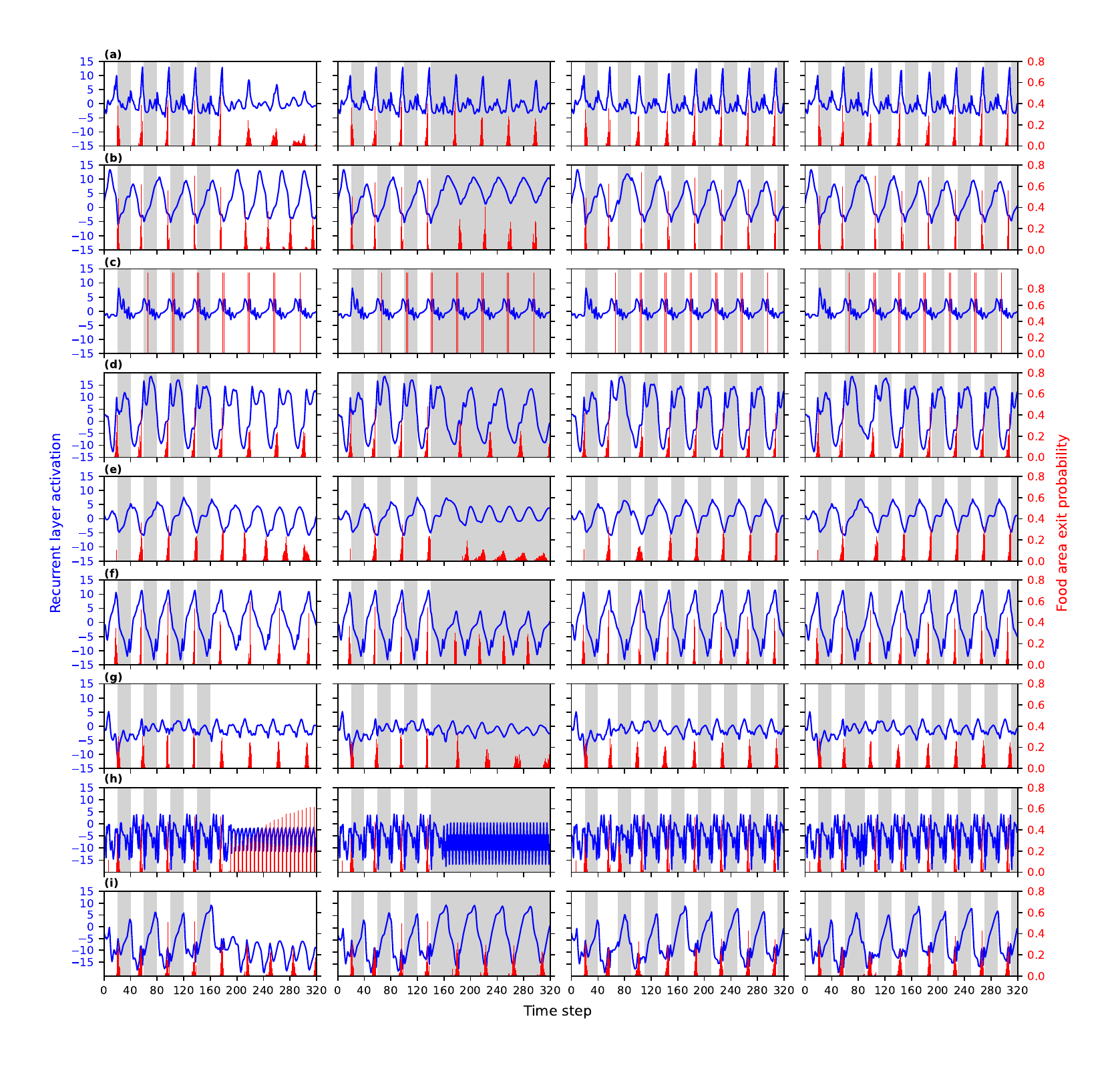}
    \caption{Effect of network and algorithmic variations on the endogeneity and entrainability of the rhythm at the recurrent layer. First and second columns display the endogeneity tests by setting a constant daylight signal from day four. Third and fourth columns display the jet-lag experiments obtained by extending the second day daytime or night by 50 \%.
    Daytime and night are represented by white and grey areas. The average activation of the neurons in the recurrent layer is plotted in blue. The red bars are histograms counting the agent's exits of the food area at the respective time step. Distribution for the food area exit timing were obtained over $1000$ test run. In training those models we used: 
    a) RMSprop optimizer,
    b) SGD optimizer,
    c) combined L1+L2 regularization on LSTM weights,
    d) He normal initialization for LSTM weights, e) LSTM layer with 32 cells, f) LSTM layer with 96 cells, g) fully connected layer with 8 units, h) vanilla RNN as recurrent layer, and i) GRU units as recurrent layer. Each case was trained over 3 or 6 million steps. For the GRU variation the exploration rate was annealed until $0.01$. Other parameters and training not mentioned in the variation remained as in the standard case (see section \ref{sec:appendix-training-details} for further training details).
    }
    \label{fig:variation1}
\end{figure}

The rows of Figure \ref{fig:variation1} (rows a-d) show the effect of the optimizer, weight regularization and initialization on the LSTM type of recurrent layer. For each case the endogeneity test are shown in the first two columns (constant day-light signals), while the entrainability (jet-lag tests) are shown in the third and fourth columns. Rows (e-i) show the effect of the type of recurrent layer and width of the recurrent and fully connected layer.

In almost all cases a sustained rhythm of the same period as the external rhythm is present even under constant daylight conditions. An exception is the case of vanilla RNN which develops a sustained rhythm of higher frequency. With respect to the jet-lag experiments, we note that in almost all cases the phase of the internal rhythm adapts to the jet lag in a two to three days inducing the agent to exit the food area at nearly the end of the day. The only exception to such adaptability occurred when a combined L1 and L2 weight regularization was applied.




\end{document}